\documentclass[review,authoryear]{elsarticle}
\usepackage{ifpdf}

\ifpdf
\usepackage[%
  pdftitle={Instructions for use of the document class
    elsart},%
  pdfauthor={Simon Pepping},%
  pdfsubject={The preprint document class elsart},%
  pdfkeywords={instructions for use, elsart, document class},%
  pdfstartview=FitH,%
  bookmarks=true,%
  bookmarksopen=true,%
  breaklinks=true,%
  colorlinks=true,%
  linkcolor=blue,anchorcolor=blue,%
  citecolor=blue,filecolor=blue,%
  menucolor=blue,pagecolor=blue,%
  urlcolor=blue]{hyperref}
\else
\usepackage[%
  breaklinks=true,%
  colorlinks=true,%
  linkcolor=blue,anchorcolor=blue,%
  citecolor=blue,filecolor=blue,%
  menucolor=blue,pagecolor=blue,%
  urlcolor=blue]{hyperref}
\fi

\makeatletter
\def\elsartstyle{%
    \def\normalsize{\@setfontsize\normalsize\@xiipt{14.5}}
    \def\small{\@setfontsize\small\@xipt{13.6}}
    \let\footnotesize=\small
    \def\large{\@setfontsize\large\@xivpt{18}}
    \def\Large{\@setfontsize\Large\@xviipt{22}}
    \skip\@mpfootins = 18\p@ \@plus 2\p@
    \normalsize
}
\@ifundefined{square}{}{\let\Box\square}
\makeatother

\usepackage{amsfonts}
\usepackage{amsmath}
\usepackage{amssymb}
\usepackage{color}
\usepackage{xcolor}
\usepackage{framed}
\colorlet{shadecolor}{blue!20}
\usepackage{graphicx}
\usepackage{setspace}

\usepackage{rotating}
\newcommand{\rv}[1]{\ensuremath{\mathbf{#1}}}
\begin{document}
\begin{frontmatter}
\title{A probabilistic framework for analysing the  compositionality of conceptual combinations}
\author[qutis]{Peter D. Bruza\corref{cor1}} 
\ead{p.bruza@qut.edu.au}
\author[qutis]{Kirsty Kitto} 
\ead{kirsty.kitto@qut.edu.au}
\author[anu]{Brentyn J. Ramm}
\ead{brentynramm@gmail.com}
\author[quteecs]{Laurianne Sitbon}
\ead{laurianne.sitbon@qut.edu.au}
\cortext[cor1]{Corresponding author}
\address[qutis]{Information Systems School\\
Queensland University of Technology\\
GPO Box 2434\\
Brisbane\\
Australia\\
tel: +61-7-31389325\\
 fax: +61-7-31384438}
\address[anu]{School of Philosophy, Australian National University}
\address[quteecs]{School of Computer Science and Electrical Engineering, Queensland University of Technology, Australia}
\begin{abstract}
Conceptual combination performs a fundamental role in creating the broad range of compound phrases utilised in everyday language. 
This article provides a novel probabilistic framework for assessing whether the semantics of conceptual combinations are compositional, and so can be considered as a function of the semantics of the constituent concepts, or not.
While the systematicity and productivity of language provide a strong argument in favor of assuming compositionality, this very assumption is still regularly questioned in both cognitive science and philosophy.
Additionally, the principle of semantic compositionality is underspecified, which means that notions of both ``strong" and ``weak" compositionality appear in the literature.
Rather than adjudicating between different grades of compositionality, the  framework presented here contributes formal methods for determining a clear dividing line between compositional and non-compositional semantics. 
In addition, we suggest that the distinction between these is  contextually sensitive.
Compositionality is equated with a joint probability distribution modeling how the constituent concepts in the combination are interpreted. 
Marginal selectivity is introduced as a pivotal probabilistic constraint for the application of the Bell/CH and CHSH systems of inequalities.
Non-compositionality is equated with a failure of marginal selectivity, or violation of either system of inequalities in the presence of marginal selectivity.
This means that the conceptual combination cannot be modeled in a joint probability distribution, the variables of which correspond to how the constituent concepts are being interpreted.
The formal analysis methods are demonstrated by applying them to an empirical illustration of twenty-four non-lexicalised conceptual combinations. 
\begin{keyword} 
conceptual combination, semantic compositionality, quantum cognition
\end{keyword}
\end{abstract} 
\end{frontmatter}

\newpage
\section{Introduction}

Humans frequently generate novel associates when presented with unfamiliar conceptual combinations.
For example, in free association experiments, subjects frequently produce the associate ``slave" when cued with the compound ``pet human" \citep{ramm:2000}, but neither ``pet" nor ``human" will have the same effect  when presented individually \citep{Nelson:McEvoy:Schreiber:2004}.
Such cases have been used by some authors to argue that conceptual combinations have a 
non-compositional semantics, as it is difficult to explain how the novel free associate ``slave" can be recovered from its constituent concepts. 

Within cognitive science, the question of how to represent even single concepts is still being debated.
Different positions have been put forward, including the prototype view, the exemplar view, and the theory theory view. \citet{murphy:2002} contrasts these positions, asking which is most supported by the various aspects of cognition related to conceptual processing, e.g., learning, induction, lexical processing and conceptual understanding in children. He concludes, somewhat disappointingly, that ``there is no clear, dominant winner''. 
Moreover, there is a well documented tension in cognitive science between the compositionality and the prototypicality of concepts, which is difficult to reconcile  \citep{frixione:lieto:2012,fodor:1998}.
Arguments in favour of  compositionality centre around the systematicity and productivity of language; there are infinitely many expressions in natural language and yet our cognitive resources are finite. Compositionality ensures that this infinity of expressions can be processed because an arbitrary expression can be understood in terms of its constituent parts.
Since compositionality is what explains systematicity and productivity,  \citet{fodor:1998} claims that concepts are, and must be compositional,
however, such claims are at odds with well-known prototypicality effects \citep{frixione:lieto:2012,fodor:1998}.
For example, consider the conceptual combination PET FISH. A GUPPY is not a prototypical PET, nor a prototypical FISH, and yet a GUPPY is a very prototypical PET FISH \citep{hampton:1997}.
Therefore, it is hard to imagine how the prototype of PET FISH can result from some composition of the prototypes of PET and FISH, which makes the characterisation of concepts in prototypical terms  difficult to reconcile with compositionality \citep{hampton:1997,fodor:1998}.
This supports a view put forward by the philosopher \citet{weiskopf:2007} when he observed that conceptual combinations are ``highly recalcitrant to compositional semantic analysis", but even this observation has garnered no general support.

Here, we approach the problem of non-compositionality from a novel perspective. We shall show that a suite of sophisticated tools have already been developed for analysing  non-compositionality, albeit  in another field of science. These tools can be naturally extended to the analysis of concepts, and provide theoretically justified grounds for deciding whether a particular conceptual combination can be considered in terms of the semantics of its constituent parts.  Specific cases will be discussed where conceptual combinations can be shown to be non-compositional using these analytical methods. 
We begin with a brief review of conceptual combination as it is currently understood in cognitive science.

\subsection{Cognitive theories, compositionality and conceptual combination}
\label{sec:theories}

The principle of compositionality states that the meanings of higher order expressions such as
sentences are determined from a combination of the meanings of their constituent parts \citep{costello:keane:2000,mitchell:lapata:2010}.
This is a principle underlying many general theories of language, both natural and artificial. 
A compositional account of conceptual combination is closely related to the notion that concepts are atomic in nature, but this assumption of atomicity is difficult to maintain when the full variety of possible semantic behavior is considered.

Perhaps most supportive of the principle are those combinations that have an intersective semantics, e.g., the meaning of BLACK CAT is the  intersection of black objects and objects that are cats.
Here, it is possible to apply a conjunction operator between the two predicates referring to  the constituent concepts, i.e., $black(x) \wedge cat(x)$.
Such intersective semantics are compositional, as the semantics of BLACK CAT are determined solely in terms of the semantics of the constituent concepts BLACK and CAT. 
It is tempting to assume that most conceptual combinations can be modeled in this way, however, the study of intersective combinations in cognitive science has revealed that not all conceptual combinations 
display such intersective semantics \citep{hampton:1997}.
For example, the intersection of ASTRONAUT and PEN in the combination ASTRONAUT PEN is empty, and therefore its semantics are vacuous, despite being a concept that humans can easily comprehend \citep{Book:00:Gardenfors:ConSpace,weiskopf:2007}. 

A second type of conceptual combination arises when the first concept modifies the head concept, e.g., in CORPORATE LAWYER, CORPORATE modifies the more general head concept to give a sub-category of LAWYER.
Schema-based theories of conceptual combination \citep{murphy:1988, wisniewski:1996}, propose that the head concept is a schema-structure made up of various property dimensions (e.g., color, size, shape etc.) and relational dimensions (e.g., habitat, functions, behaviors etc.). 
Several studies have revealed that modification can produce emergent properties, e.g., in HELICOPTER BLANKET the modification of BLANKET by HELICOPTER generates associate properties such as ``water proof", ``camouflage", and ``made of canvas", a phenomenon which present theories struggle to account for \citep{wilkenfeld:ward:2001}, and so is sometimes viewed as evidence for non-compositional semantics \citep{hampton:1997,medin:shoben:1988}. 

Despite these tensions underlying the assumption of compositionality, virtually all researchers have at least assumed a weak form of compositionality in their analysis of human language, where for example, the initial combination process begins with separate meanings, but is supplemented later by external contextual information \citep{wisniewski:1996,swinney:love:walenski:smith:2007}.
For example, in \citet{wisniewski:1996}'s dual process theory of conceptual combination, a competition occurs between the processes of relation linking (e.g., ZEBRA CROSSING as a crossing for zebras), and property mapping (e.g., ZEBRA CROSSING as a striped crossing), as the meaning of the compound is decided upon. 
This process is affected by the similarity of the constituent concepts, because similar concepts share many facets and so are more likely to result in a property interpretation, whereas dissimilar concepts are more likely to be combined using a relational process. Thus,  ELEPHANT HORSE is more likely to result in a property interpretation (e.g., a large horse), than ELEPHANT BOX, which is more likely to result in a relational interpretation (e.g., a box for holding elephants), because similar concepts share many dimensions (four legs, similar shape etc. in the case of elephant and horse) and thus are easier to combine by mapping one property to another. However, it is important to note that these processes are all weakly compositional, in the sense that they rely almost exclusively on the properties of the individual concepts. It is only later that background knowledge is drawn upon to infer the possible emergent properties of the new concept. Thus an ELEPHANT BOX could be deemed as likely to be made of a 
strong material such as wood, and hopefully to contain air-holes.
\citet{swinney:love:walenski:smith:2007} found  evidence for this form of weak compositionality in conceptual combination,  when they showed that for adjectival combinations such as BOILED CELERY the properties of the individual words such as ``green" are activated before emergent properties such as ``soft". 
However, for the combination APARTMENT DOG,  apartment modifies the ``habitat" dimension of dog rather than its ``size" (a dog the size of an apartment), which in turn shows that background knowledge also plays a role in early combinatory processes such as slot selection \citep{murphy:1988}. 

Rather than entering the debate about the proper dividing line between weak and strong compositionality, it is our intention to provide a formal framework to analyse the (non-)compositionality of conceptual combinations, motivated by the analysis of composite systems in quantum physics. An important point is that this framework can be empirically tested. Thus, we feel that it is possible to shift this debate out of philosophy and into the realms of experimental psychology,\footnote{In much the same way as the field of physics entered the realms of experimental testing with the work of Bell and Aspect, after decades of more philosophical debate as to the separability and completeness of the quantum formalism \citep{isham:1995,laloe:2001}.} and this article is a step in that direction.
In what follows we shall discuss the combination of concepts within a tiered model of cognition. This will provide a framework from which a (non-)compositional semantics can be developed in further sections.

\section{Probabilistic approaches to modeling conceptual combinations}

It is at the symbolic level of cognition where  a significant portion of the work on compositional semantics can be placed because this is where higher order symbolic structures and associated rules, such as grammar, are processed.
A grammar specifies the parts of a sentence, and the manner in which they fit together. 
It makes sense that the semantics attributed to these primitive parts be intuitive, for example, a  noun may be mapped to a set of entities.
However, \citet{Zadrozny:1992} has suggested that it doesn't actually matter which components are chosen as primitive, a function can  be found that will \emph{always} produce a compositional semantics. In Zadrozny's own words, ``..compositionality, as commonly defined, is not a strong [enough] constraint on a semantic theory''. 
The consequence of this with respect to the compositional semantics of natural language, and hence conceptual combination, is that meaning need not be assigned to individual words, ``we can do equally well by assigning meaning to phonemes or even LETTERS. . . " \citep{Zadrozny:1992}.  
Opponents to Zadrozny may argue that his position is overly pessimistic because it applies to ``strong compositionality", a position that is clearly wrong. 
Nevertheless, the question remains as to where the ``meanings" might come from initially.

Consider the concept BAT. 
One reliable way to seek an understanding of this concept  is via free association experiments where subjects are cued with the word ``bat" and asked to produce the first word that comes to mind. 
Over large numbers of subjects, probabilities can be calculated that a certain associate is produced. Fig.~\ref{boxer-bat-free}(a) depicts such a set of data taken from the University of South Florida word association norms (USF-norms) \citep{Nelson:McEvoy:Schreiber:2004}.
Upon examination of this table, we can see that these probabilities represent two clear senses for the cue ``bat''; a SPORT sense (with relevant associates in bold) and an ANIMAL sense. Considering the full dataset\footnote{Available at http://web.usf.edu/FreeAssociation/AppendixC/Matrices.A-B .} allows us to generate the total probability $p_{s}$ of recall for the sport sense by summing the probabilities of the relevant associates: 
$p_{s}= 0.25 + 0.05 = 0.30$.
The rest of the associates all happen to be relevant to the animal sense of bat, so $p_{a} = 0.70$.
\begin{figure}
\begin{minipage}{0.45\linewidth}
\centering
{\small 
\begin{tabular}{|l|r|} \hline 
{\em Associate} & {\em Probability} \\ \hline
{\bf ball} & {\bf 0.25} \\
cave & 0.13 \\
vampire & 0.07 \\
fly & 0.06 \\
night & 0.06 \\
{\bf baseball} & {\bf 0.05} \\
bird & 0.04 \\
blind & 0.04 \\
animal & 0.02 \\
$\cdots$ & $\cdots$ \\ \hline
\end{tabular}}\\\vspace*{0.5cm}
(a)
\end{minipage}
\begin{minipage}{0.45\linewidth}
\centering
{\small
\begin{tabular}{|l|r|} \hline 
{\em Associate} & {\em Probability} \\ \hline
{\bf fighter} & {\bf 0.14} \\
{\bf gloves} &{\bf 0.14} \\
{\bf fight} & {\bf 0.09} \\
dog & 0.08 \\
{\bf shorts} & {\bf 0.07} \\
{\bf punch} & {\bf 0.05} \\
{\bf Tyson} & {\bf 0.05} \\
$\cdots$ & $\cdots$ \\ \hline
\end{tabular}}\\\vspace*{0.5cm}
(b)
\end{minipage}
\caption{(a) Free association probabilities for the word ``boxer" (a) and the word ``bat" (b).}
\label{boxer-bat-free}
\end{figure}
The same can be said for the concept BOXER (see Fig.~\ref{boxer-bat-free}(b) where, once again, the associates relevant to the sport sense of BOXER are in bold). 

When the preceding is considered in relation to how the conceptual combination BOXER BAT may be interpreted, then four interpretations are possible.
For example, when BOXER is interpreted as a sport and BAT as an animal, 
the corresponding interpretation of the combination maybe something along the lines of a ``furry black animal with boxing gloves on", or perhaps BOXER could be interpreted as a sport and BAT as as a sport leading a subject to interpret the compound as  ``a fighter's implement".

Conceptual combinations usually have more than one possible interpretation. 
This may arise from a range of factors, including the meaning of the concepts themselves (e.g. BOXER can be interpreted as a dog, a sportsperson, a pair of shorts, someone who puts things in boxes, etc.), the sentence in which they appear, the background of the subject etc..
Different human subjects will often interpret the same conceptual combination differently, indeed, the same human subject, if placed in a new context may very well provide a different interpretation for the same concept. Thus, it is sensible to approach the analysis of compositionality probabilistically. 

In what follows each concept is assumed to have a \emph{dominant} sense and one or more  \emph{subordinate} senses. The distinction between the two can be inferred from free association norms such as those discussed above. For example, the dominance of the sport sense of BOXER is clearly evident in Fig.~\ref{boxer-bat-free}(a), where the probability associated with the sport sense is greater than the animal sense,  which leads us to designate the sport sense as ``dominant" and the ``animal" sense as subordinate. 
It should be noted, however, that the distinction between ``dominant" and ``subordinate" senses is not necessary for the theory presented below, rather it is an explanatory aid.

Standard probabilistic reasoning suggests that if two ambiguous  concepts $A$ and $B$ have behavior that can be considered as compositional, then it should be possible to describe this behavior in terms of four dichotomous random variables, $\{\rv{A1,A2}\}$ and $\{\rv{B1,B2}\}$, ranging over two values $\{+1,-1\}$. 
The numbers 1 (dominant) and 2 (subordinate) correspond to the  senses attributed to the respective concepts $A$ and $B$. 
However, if a human subject is first shown the word ``vampire" and subsequently asked to interpret the compound BOXER BAT, then they may be oriented towards giving an animal interpretation of BAT. 
This suggests a  minimal natural extension where $\rv{A1}=+1$ represents a situation where the dominant sense of concept $A$ was first primed  and concept $A$ was indeed subsequently interpreted in that sense by the human subject.
Conversely, $\rv{A1} = -1$ represents the case where the dominant sense of concept $A$ was primed  but $A$ was not interpreted in that sense. 
Similarly, $\rv{A2}=+1$ represents a situation where a subordinate sense of concept $A$ was first primed, and concept $A$ was indeed subsequently interpreted in this sense, and
$\rv{A2} = -1$ represents the case where a subordinate sense of concept $A$ was primed, but $A$ was not interpreted in this subordinate sense.
Note that a concept may have more than one subordinate sense. 
For example, the concept BOXER could be considered to have subordinate clothing sense, namely ``boxer shorts".
Therefore, the previous specification based on a primary and subordinate sense does not preclude that the concept $A$ is interpreted in a third sense, for example, when $\rv{A1} = -1$, this can occur when concept is interpreted in the subordinate sense modeled by $\rv{A2}$, or a third sense. 
Similar relationships hold for $\rv{B1}$ and $\rv{B2}$.

Priming thus allows for the experimental control of the contextual cues influencing conceptual combinations. This is important because conceptual combinations always appear in a context (e.g., a discourse context), which affects how they will ultimately be interpreted.  
\begin{figure}[h]
\centering
 \includegraphics[width=12cm]{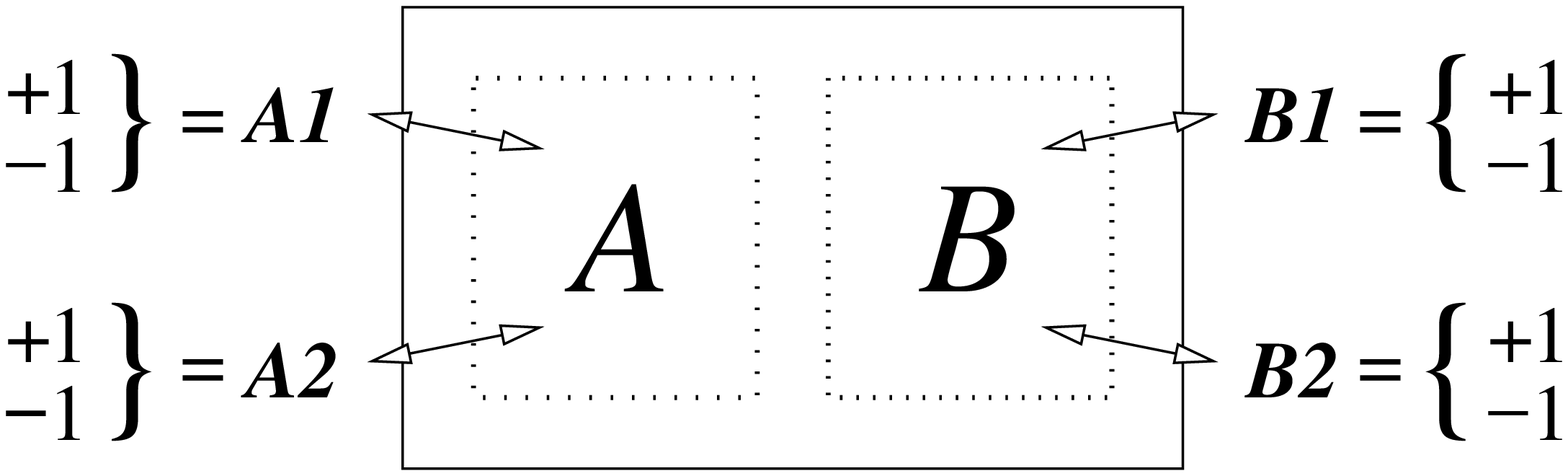}
\caption{A potentially compositional system $S$, consisting of two assumed components $A$ and $B$. $S$ can perhaps be understood in terms of a mutually exclusive choice of experiments performed upon those components, one represented by the random variables $\rv{A1},\rv{A2}$ (pertaining to an interaction between the experimenter and component $A$), and the other by $\rv{B1},\rv{B2}$ (pertaining to an interaction between the experimenter and component $B$). Each of these experiments can return a value of +1 or -1}\label{fig:system}
\end{figure}
Fig.~\ref{fig:system} gives a general representation of the reasoning used in the construction of the above probabilistic scenario. A `black box' is depicted, with two proposed components, $A$ and $B$, inside it. Two different experiments can be carried out upon each of the two presumed components, which will answer a set of `questions' with binary outcomes, leading to four experimental scenarios. 
For example, one experimental scenario would be to ask whether subjects return an interpretation of the concept $\rv{A}$ that corresponds to the prime $\rv{A1}$ and similarly for $\rv{B}$ in relation to the prime $\rv{B2}$.
What analysis can be brought to bear upon such a situation? 

As with many systems, the outcomes of our experiments will have a statistical distribution over all available outcomes. In what follows, we shall aim to develop a general mathematical apparatus that can be used to discover whether the presumed sub-components can be considered as isolated, influencing one another, or in some sense irreducible. We shall do this through a consideration of the joint probability distribution $\Pr(\rv{A1},\rv{A2},\rv{B1},\rv{B2})$ which is used to model the behavior of the experimental black box. 
While this analysis will be performed using conceptual combinations, we emphasise that this black box is potentially very general and that the analysis developed here can be applied to far more than the analysis of language. 

We start by noting that we can construct 16 joint probabilities, corresponding to all the possible interpretations of concepts $A$ and $B$ that a subject might return, across the four priming conditions:
\begin{align}
\label{eq:ps}
 p_1  \equiv \Pr(\rv{A1}=+1,\rv{B1}=+1) \qquad\qquad p_2 &\equiv \Pr(\rv{A1}=+1,\rv{B1}=-1) \nonumber\\
 p_3 \equiv \Pr(\rv{A1}=-1,\rv{B1}=+1) \qquad\qquad p_4 &\equiv \Pr(\rv{A1}=-1,\rv{B1}=-1)  \nonumber\\
p_5 \equiv \Pr(\rv{A1}=+1,\rv{B2}=+1) \qquad\qquad p_6 &\equiv \Pr(\rv{A1}=+1,\rv{B2}=-1) \nonumber\\
p_7 \equiv \Pr(\rv{A1}=-1,\rv{B2}=+1) \qquad\qquad p_8 &\equiv \Pr(\rv{A1}=-1,\rv{B2}=+1)  \nonumber\\
p_9 \equiv \Pr(\rv{A2}=+1,\rv{B1}=+1) \qquad\qquad p_{10} &\equiv \Pr(\rv{A2}=+1,\rv{B1}=-1)  \nonumber \\
p_{11} \equiv \Pr(\rv{A2}=-1,\rv{B1}=+1) \qquad\qquad p_{12} &\equiv \Pr(\rv{A2}=-1,\rv{B1}=-1)  \nonumber \\
p_{13} \equiv \Pr(\rv{A2}=+1,\rv{B2}=+1) \qquad\qquad p_{14} &\equiv \Pr(\rv{A2}=+1,\rv{B2}=-1)  \nonumber \\
p_{15} \equiv \Pr(\rv{A2}=-1,\rv{B2}=+1) \qquad\qquad p_{16} &\equiv \Pr(\rv{A2}=-1,\rv{B2}=-1).
\end{align}
These sixteen probabilities can be set out in an array as follows:
\begin{equation}
\begin{array}{c}
A%
\end{array}%
\begin{array}{c}
\underset{}{%
\begin{array}{c}
\rv{A1}%
\end{array}%
\begin{array}{c}
+1 \\ 
-1%
\end{array}%
} \\ 
\overset{}{%
\begin{array}{c}
\rv{A2}%
\end{array}%
\begin{array}{c}
+1 \\ 
-1%
\end{array}%
}%
\end{array}%
\overset{\overset{%
\begin{array}{c}
B%
\end{array}%
}{%
\begin{array}{cc}
\overset{%
\begin{array}{c}
\rv{B1}%
\end{array}%
}{%
\begin{array}{cc}
+ 1\ \  & -1\qquad%
\end{array}%
} & \overset{%
\begin{array}{c}
\rv{B2}%
\end{array}%
}{%
\begin{array}{cc}
+1\ \  & -1%
\end{array}%
}%
\end{array}%
}}{\left( 
\begin{tabular}{c|c}
$\underset{}{%
\begin{array}{cc}
p_{1} & p_{2} \\ 
p_{3} & p_{4}%
\end{array}%
}$ & $\underset{}{%
\begin{array}{cc}
p_{5} & p_{6} \\ 
p_{7} & p_{8}%
\end{array}%
}$ \\ \hline
$\overset{}{%
\begin{array}{cc}
p_{9} & p_{10} \\ 
p_{11} & p_{12}%
\end{array}%
}$ & $\overset{}{%
\begin{array}{cc}
p_{13} & p_{14} \\ 
p_{15} & p_{16}%
\end{array}%
}$%
\end{tabular}%
\right) }  = P_{AB} \label{pstruct}
\end{equation}
This matrix lists the different priming conditions in a set of four blocks, which allows us to consider the structure of the probabilities describing the likely interpretation of a given conceptual combination.
Observe how the matrix $P_{AB}$ in equation \eqref{pstruct} is complete, in that it covers all possible priming conditions across the respective senses of the concepts. 

In what follows we will show that $P_{AB}$ can be used to determine whether a conceptual combination is compositional, or not. We start by considering what might be required in order for a conceptual combination to be deemed compositional.

\subsection{Compositional semantics}
\label{sec:compositional}

Were the semantics of the conceptual combination $AB$ to be compositional, how would this be reflected in its probabilistic structure?
The principle of semantic compositionality would suggest that the joint probability distribution could be recovered from the probability distributions constructed using each individual concept.
For example, a naive assumption would be that the concepts in the combination can be interpreted independently of one another, 
\begin{align}
\Pr(\rv{Ai},\rv{Bj})&= \Pr(\rv{Ai})\Pr(\rv{Bj}), i,j \in \{1,2\}. \label{eqn:naive}
\end{align}
The syntax of this equation clearly reflects how the probabilistic behaviour of the conceptual combination is represented by the four joint distributions $\Pr(\rv{Ai},\rv{Bj})$.

Less naive formalisations of compositionality are possible. 
For example, the competition among relations in nominals (CARIN) theory of conceptual combination proposes sixteen possible relations that can be used to link concepts, e.g., \emph{causes}, \emph{during}, \emph{for} and \emph{about} \citep{gagne:shoben:1997,gagne:2001}. 
One possible assumption assumes that relation independently affects the interpretation of each concept:
\begin{align}
\Pr(\rv{Ai},\rv{Bj}| \rv{R})&=  \Pr(\rv{Ai}|\rv{R})\Pr(\rv{Bj}|\rv{R}), i,j \in \{1,2\}. \label{eqn:CH74}
\end{align}
where $\rv{R}$ is a random variable over the  possible implicit linking relations.
CARIN, however, assumes that when people interpret a novel conceptual combination $AB$, the availability of each of these sixteen relations is determined by the relation frequency distribution for the modifying concept $A$. The random variable $\rv{R}$ is thus assumed to range over these sixteen relations.
Therefore, CARIN explains why MOUNTAIN GOAT is easier to interpret than MOUNTAIN MAGAZINE because the \emph{located} relation is more often used with the modifier MOUNTAIN than the \emph{about} relation. 
The essence of the CARIN approach to interpreting conceptual combinations can therefore be derived as follows using using \eqref{eqn:CH74} together with the assumption that the interpretation of concept $B$ is independent of the linking relation,
\begin{align}
\Pr(\rv{Ai},\rv{Bj})  &= \sum_{r\in \rv{R}}\Pr(\rv{Ai},\rv{Bj}|r)\Pr(r) \\ 
                                &= \sum_{r\in \rv{R}}\Pr(\rv{Ai}|r)\Pr(\rv{Bj}|r)\Pr(r) \\
                                &= \sum_{r\in \rv{R}}\Pr(\rv{Ai},r)\Pr(\rv{Bj})  \label{eqn:CARIN}
\end{align}
for $i,j\in\{1,2\}$. 
In other words, a formalisation of the CARIN model is as follows which shows how the the relation frequency distribution for the modifying concept $A$ is formalised by the joint distribution $\Pr(\rv{Ai},\rv{R})$.
The goal of the preceding development of equations is not to formalise existing models, or propose new ones, but to introduce how compositionality may defined in a probabilistic way.
Observe that the equations \eqref{eqn:naive}\eqref{eqn:CH74},\eqref{eqn:CARIN} all define four pairwise joint probability distributions $\Pr(\rv{Ai},\rv{Bj}),i,j \in \{1,2\}$.

A given conceptual combination AB is deemed \emph{compositional} if and only if a four way joint distribution 
$\Pr(\rv{A1},\rv{A2},\rv{B1},\rv{B2})$ exists where $\Pr(\rv{Ai},\rv{Bj}),i,j \in \{1,2\}$ are marginal distributions.
This opens the door to define non-compositionality via an unusual means, namely the inability to construct a joint probability distribution $\Pr(\rv{A1},\rv{A2},\rv{B1},\rv{B2})$ in this way.

\subsection{Non-compositional semantics}
To analyse non-compositionality we draw upon results from the field of quantum theory in relation to entangled systems.
This step is not as arbitrary as it might at first seems.
An entangled system is one for which is it impossible to construct a four way joint distribution from four empirically collected pairwise joint distributions.
Fine's theorem is one of the results from quantum theory which allows entanglement to be formally defined in this way.
Importantly, the theorem states the necessary and sufficient conditions for  the notion compositionality introduced at the end of the previous section.


\citet{Fine:1982} provides both the \emph{necessary and sufficient conditions}  for the existence of the joint probability distribution $\Pr(\rv{A1},\rv{A2},\rv{B1},\rv{B2})$:

\emph{{\bf  Fine Theorem 3 \citep{Fine:1982}:}} If $\rv{A1},\rv{A2},\rv{B1},\rv{B2}$ are bivalent random variables with joint distributions $\Pr(\rv{Ai},\rv{Bj}),\ i,j \in \{1,2\}$, then necessary and sufficient for a joint distribution $\Pr(\rv{A1},\rv{A2},\rv{B1},\rv{B2})$ is that the following system of inequalities is satisfied:
\begin{small}
\begin{align}
-1 \leq \Pr(A1,B1) + \Pr(A1,B2) + \Pr(A2,B2) - \Pr(A2,B1) - \Pr(A1) - \Pr(B2) \leq 0 \label{bch1}\\
-1 \leq \Pr(A2,B1) + \Pr(A2,B2) + \Pr(A1,B2) - \Pr(A1,B1) - \Pr(A2) - \Pr(B2) \leq 0 \label{bch2}\\
-1 \leq \Pr(A1,B2) + \Pr(A1,B1) + \Pr(A2,B1) - \Pr(A2,B2) - \Pr(A1) - \Pr(B1) \leq 0 \label{bch3}\\
-1 \leq \Pr(A2,B2) + \Pr(A2,B1) + \Pr(A1,B1) - \Pr(A1,B2) - \Pr(A2) - \Pr(B1) \leq 0. \label{bch4}
\end{align}
\end{small}
Fine referred to this system of inequalities as the Bell/CH inequalities, and we will adhere to his labeling. 
Fine's theorem permits us to analyse compositionality from the perspective of the above four equations,
namely, a conceptual combination AB is deemed ``non-compositional" when the four pair wise joint probability distributions depicted in table \eqref{pstruct} do not satisfy the Bell/CH inequalities.
This implies then a joint distribution $\Pr(\rv{A1},\rv{A2},\rv{B1},\rv{B2})$ \emph{cannot} be formed such that the four pairwise joint probability distributions $\Pr(\rv{Ai},\rv{Bj}),i,j \in \{1,2\}$ are marginal distributions. 
Conversely, if all inequalities are satisfied the four way joint probability distrubution does exist and the conceptal combination is thus deemed ``compositional".

In quantum physics, entangled systems adhere to a constraint variously termed the ``causal communication constraint", ``parameter independence", ``simple locality", ``signal locality", or ``physical locality".  
This constraint expresses ``the probability of a particular measurement outcome on any one part of the system should be independent of which sort of measurement was performed on the other parts" \citep{cereceda:2000}. 
In the context of cognitive science, this constraint has been termed ``marginal selectivity" \citep{dzhafarov:kujala:2012}.
For example, with respect to the conceptual combination BOXER BAT, marginal selectivity entails the interpretation of BAT does not change when the primes of BOXER are varied from ``fighter" to ``dog".
Marginal selectivity is expressed more formally as follows:
\begin{align}
\Pr(Ai) &= \Pr(Ai,B1)+ \Pr(Ai,\bar{B1}) = \Pr(Ai,B2)+ \Pr(Ai,\bar{B2}), i \in \{1,2\} \\
\Pr(Bj) &= \Pr(A1,Bj)+ \Pr(\bar{A1},Bj) = Pr(A2,Bj)+ \Pr(\bar{A2},Bj), j \in \{1,2\}
\end{align}
Note how these four equations express that the interpretation of the concept represented by the marginal probability, e.g., $\Pr(Ai), i \in \{1,2\}$  is stable with respect to how the other concept is primed, represented by $B1$ and $B2$.

Recently, \citet{dzhafarov:kujala:2012} have established a connection between cognitive modeling and Fine's theorem using the theory of selective influences, a result that adds to the cognitive validity of Fine's theorem.
In a model with several factors  and a set of random variables describing responses, \emph{selective influence} concerns the problem of what factors influence what variables. 
The interpretation of conceptual combinations within  a priming scenario can be treated with a model of selective influence, with primes corresponding to the factors affecting random variable corresponding to the interpretation of concepts.
\citet{dzhafarov:kujala:2012} point out that selective influence implies marginal selectivity. 
Failure of marginal selectivity means there can be no model of selective influence, meaning there is no joint probability distribution $\Pr(\rv{A1},\rv{A2},\rv{B1},\rv{B2})$ where the pairwise 
distributions $\Pr(\rv{A1},\rv{B1})$, $\Pr(\rv{A1},\rv{B2})$, $\Pr(\rv{A2},\rv{B1})$, $\Pr(\rv{A2},\rv{B2})$ are marginal distributions.

The proof of Fine's theorem assumes marginal selectivity, a constraint that holds for entangled systems of photons in quantum physics. 
In cognitive science, however, concepts are not as well behaved as photons, so marginal selectivity may or may not hold. 
This is a crucial point.
For applications in cognitive science, marginal selectivity must first be tested before Fine's theorem can be applied:
\begin{enumerate} 
\item If marginal selectivity fails, then the conceptual combination is \emph{immediately}  judged as ``non-compositional".
\item If marginal selectivity holds and any of the Bell/CH inequalities are violated, then the conceptual combination is deemed ``non-compositional".
\item If marginal selectivity holds and all of the Bell/CH inequalities ahold, then the conceptual combination is deemed ``compositional".
\end{enumerate}

In quantum physics, the so called CHSH inequality has also been used to analyse entangled systems. 
The advantage of the CHSH inequality is that its formulation based on correlations permits some insight to be gained into why conceptual combinations are (non-)compositional.
The CHSH inequality is as follows  \citep{cereceda:2000}:
\begin{eqnarray}
-2 \leq E(\rv{A1},\rv{B1}) + E(\rv{A1},\rv{B2}) + E(\rv{A2},\rv{B1}) - E(\rv{A2},\rv{B2})   \leq 2 \label{eq:cereceda}
\end{eqnarray}
The expectations can easily be computed from the matrix of probabilities \eqref{pstruct}.
For example, $E(\rv{A1},\rv{B1}) = p1 + p4 - (p2 + p3)$.
Recalling from \eqref{eq:ps} that $p_1 = \Pr(\rv{A1}=+1,\rv{B1}=+1)$ and    $p_4 = \Pr(\rv{A1}=-1,\rv{B1}=-1)$, we recognize that $p_1$ corresponds to a situation where concepts $A$ and $B$ have \emph{both} been interpreted in their dominant sense, when in both cases the dominant sense of each concept has been primed.
Similarly, $p_4$ corresponds to both $A$ and $B$ being interpreted in a subordinate sense when the dominant sense of each concept has been primed. 
Thus, $p_1+p_4=1$ occurs when the senses of the constituent concepts are perfectly correlated within the given priming condition. 
For example, assuming that the fruit sense of APPLE was primed and food sense of CHIP was primed, perfect correlation of senses in this priming condition means two conditions hold: (1)  when APPLE is interpreted as a fruit CHIP is always interpreted as food ($p1$) and (2)  when APPLE is \emph{not} interpreted as fruit, CHIP is \emph{not} interpreted as food ($p4$).
The combination of these two conditions imply that $p_1+p_4=1$ and $p_2 + p_3=0$. 
Conversely, $p_2+p_3 = 1$ occurs when the senses are perfectly anti-correlated. 
For example, assume  the fruit sense of APPLE is primed and CHIP is primed in its electronic circuit sense.
Perfect anti-correlation of senses means two conditions hold: (1)  when APPLE is \emph{not} interpreted as a fruit, CHIP is always interpreted as a circuit ($p3$) and (2)  when APPLE is interpreted as fruit, CHIP is \emph{not} interpreted in its circuit sense ($p2$).

With this as underlying intuition, the expectation value $E(\rv{Ai},\rv{Bj})$ captures how well the senses of the constituent concepts are (anti-)correlating.
The arrangement of probabilities in figure \eqref{pstruct} is not significant. There are thus four possible ways to arrange the quadrants, each arrangement leading to a variant of the CHSH inequality:  
\begin{eqnarray}
-2 \leq E(\rv{A1},\rv{B1}) - E(\rv{A1},\rv{B2}) + E(\rv{A2},\rv{B1}) + E(\rv{A2},\rv{B2})   \leq 2 \\
-2 \leq E(\rv{A1},\rv{B1}) + E(\rv{A1},\rv{B2}) - E(\rv{A2},\rv{B1}) + E(\rv{A2},\rv{B2})   \leq 2 \\
-2 \leq -E(\rv{A1},\rv{B1}) + E(\rv{A1},\rv{B2}) + E(\rv{A2},\rv{B1}) + E(\rv{A2},\rv{B2})   \leq 2 \\
\label{eq:cereceda}
\end{eqnarray}
Therefore, there are four CHSH inequalities, each differing where the minus sign is placed.
The heart of each inequality is a computation involving expectations which will be referred to as the CHSH value.
When the CHSH value of any of the inequalities lies outside of the range [-2, 2], meaning its absolute value is greater than 2, there is no joint probability distribution $\Pr(\rv{A1}, \rv{A2}, \rv{B1}, \rv{B2})$ such that the four empirically collected pairwise distributions 
$\Pr(\rv{A1}, \rv{B1}), \Pr(\rv{A1}, \rv{B2}), \linebreak
\Pr(\rv{A2}, \rv{B1}), \Pr(\rv{A2}, \rv{B2})$ are marginal distributions.
In such a case, the associated conceptual combination is deemed ``non-compositional"

Conversely, when the CHSH value lies within [-2,2] for all four inequalities, there \emph{is} a joint probability distribution $\Pr(\rv{A1}, \rv{A2}, \rv{B1}, \rv{B2})$ where the four empirically collected pairwise distributions: $\Pr(\rv{A1}, \rv{B1}), \Pr(\rv{A1}, \rv{B2}),
 \Pr(\rv{A2}, \rv{B1}), \linebreak
 \Pr(\rv{A2}, \rv{B2})$ are marginal distributions.
In this case, the conceptual combination is deemed ``compositional".

As was the case with Bell/CH inequalities, marginal selectivity must first be tested before the four CHSH inequalities can be applied.
The formal connection between the CHSH inequalities and Fine's theorem is as follows: By assuming marginal selectivity together with the CHSH inequalities, the Bell/CH inequalities can be derived.

In summary, both the Bell/CH inequalities and the CHSH inequalities allow non-compositionality to be determined by the inability to construct a joint probability distributions across the four variables modeling how the primary and a subordinate sense of the concepts $A$ and $B$ are interpreted.
We now illustrate how these probabilistic methods for analysing compositionality can be deployed in an experimental setting.

\section{Empirical Illustration}
\label{sec:empirical}
\subsection{Subjects}
Sixty-five subjects were recruited from the undergraduate psychology pool at Griffith University and received credit for their participation.
Only native English speakers were selected in order remove the possibility that the interpretation of conceptual combinations would be confounded by language issues.

\subsection{Design and materials}

We utilised four different priming regimes in order to generate the four different experimental scenarios suggested by Fig.~\ref{fig:system}. 
In these experiments, subjects were first primed and then presented with a non-lexicalised conceptual combination,
which they were asked to interpret also designating the senses that were used in that interpretation.
A probabilistic analysis was then performed upon the data so obtained. Subjects were presented with twenty-four `true' conceptual combinations (see below for an explanation), and so participated in twenty-four test trials.
Table \ref{tab:results} lists the set of conceptual combinations used, as well as  the corresponding primes.  

Primes were selected from the USF free association norms \citep{Nelson:McEvoy:Schreiber:2004} and the University of Alberta norms of homographs \citep{twilley:dixon:taylor:clark:1994}. 
The majority of primes were selected from the USF norms.
The procedure for selecting primes from these norms was to view a potential prime as a cue which produces the concept as an associate.
As an example, ``money"  was chosen from the USF norms to prime the financial sense of BANK as ``bank" is produced as a free associate of the cue `money'.
Similarly, ``river" was chosen to prime the natural sense of BANK. 
Occasionally when a particular sense was not present in the USF norms,  we drew upon the University of Alberta norms. 
Importantly,  the USF norms were used to avoid cues such as `accountÕ which was associated with both BANK  and LOG, thereby minimising the possibility of priming more than one concept at a time. 

A single factor design was used, which analysed responses to non-lexicalised conceptual combinations under priming conditions that varied between subjects. 
A subject was assigned to one of four priming conditions for each presented conceptual combination. 
For example, the four priming conditions for BANK LOG are (1) ``money'' and ``journal'' ($\rv{A1}-\rv{B1}$), (2) ``money'' and ``tree'' ($\rv{A1}-\rv{B2}$), (3) ``river'' and ``journal'' ($\rv{A2}-\rv{B1}$), or (4) ``river'' and ``tree'' ($\rv{A2}-\rv{B2}$). 
This assignment of primes was based upon a between groups Latin square design, such that for the 24 combinations, each participant completed each priming condition 6 times. 
Combinations were chosen with the expectation that the ambiguity of its constituents would allow a number of alternative interpretations, where each interpretation arose from a different attribution of meaning to the underlying sense of the ambiguous concepts \citep{costello:keane:1997a}.

\subsection{Procedure}
\begin{figure}
\centering
\includegraphics[width=12cm]{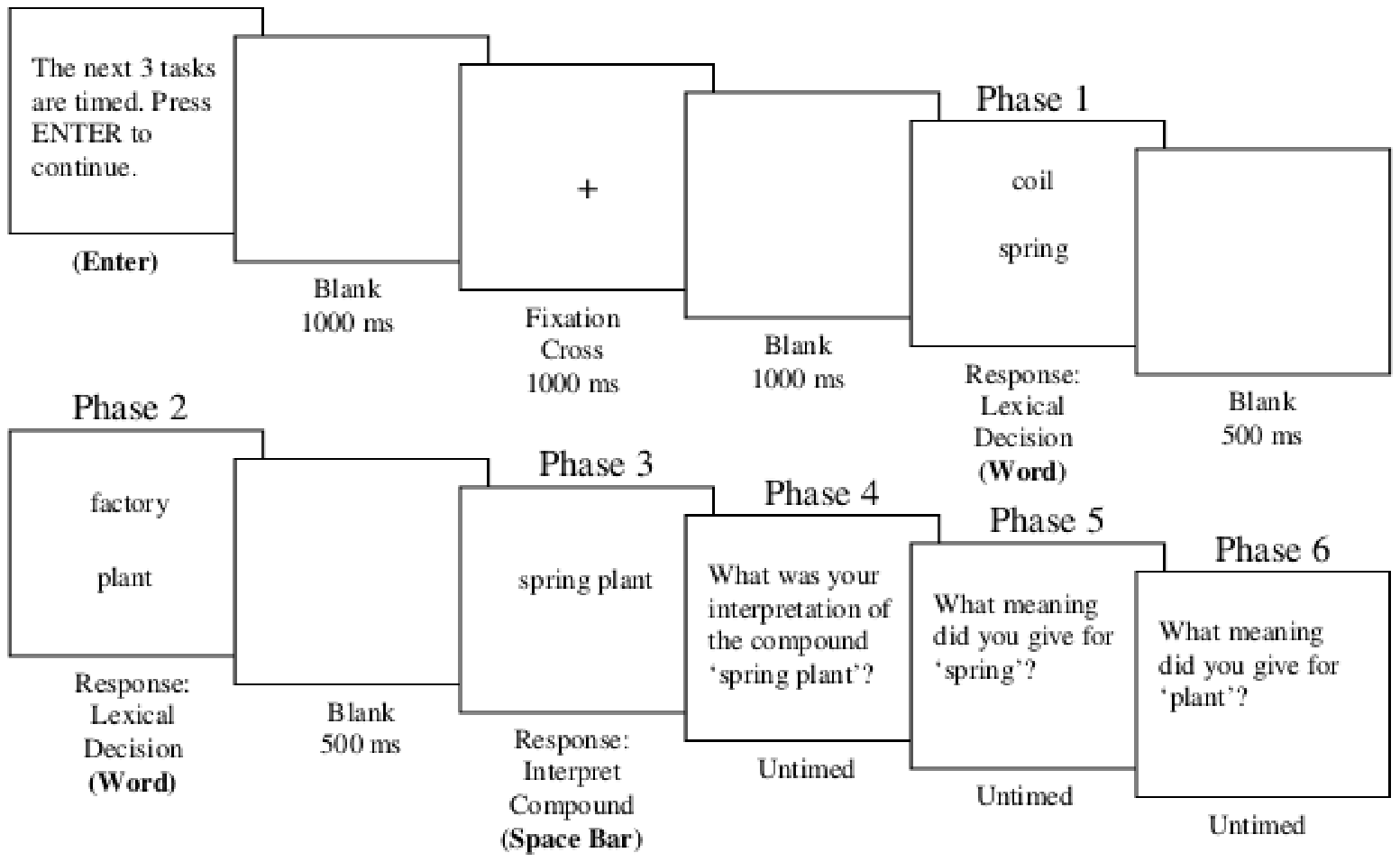}
\caption{Example experimental structure for a trial. Non-word trials followed a similar structure, with primes in Phase 1 and/or Phase 2 replaced with non-words. The sequence of squares moving from left to right show the experimental flow, with each square a representation of the screen shown to a participant. Note: the figure does not show the exact text given to participants, and stimuli are not to scale.} \label{fig:trialStructure}
\end{figure}

Participants completed 3 practice trials, 24 test trials and 24 filler
trials, and Fig.~\ref{fig:trialStructure} shows a schematic illustration of the procedure followed during a test trial.  All trials were composed of six phases, consisting of three initial
time-pressured tasks followed by three non-timed tasks. The time limitation of the first three phases was utilised in order to maximise the effectiveness of the priming. 
The experiment took around 20--30 minutes to complete, and participants pushed the ENTER key to begin each trial. 

\noindent
{\bf Phases 1-2:}\\
Two consecutive double lexical decision tasks were carried out, where participants were asked to decide as quickly as possible whether two letter strings, a prime and the concept to be presented as a part of the compound given in Phase 3, were legitimate words, or if one of the strings was a non-word. 
Each lexical decision consisted of the the two letter strings presented in the center of screen, one below the other, in order to discourage participants from interpreting the two words as a phrase. Participants responded to the decision tasks by pushing a button on the keyboard, labeled `word' or a button labeled `non-word' (left arrow and right arrow keys respectively).  For instance, if given the strings ``coil'' and ``spring'', then participants were expected to decide that both strings were words and so push the `word' key, whereas if given  ``grod'' and ``church'' then participants were expected to decide that they had been shown a non-word and to push the `non-word' key. 
For all of the test trials participants received two phases of word-word strings. The response ratio for the two priming phases were: 50\% word $\rightarrow$ word (test trial), 25\% non-word $\rightarrow$ non-word (filler trial), 12.5\% word $\rightarrow$ non-word (filler trial), 12.5\% non-word $\rightarrow$ word (filler trial). In phases where a non-word was present, it appeared equally often in the top or the bottom portion of the screen.  

The double-lexical decision task was used to associate the priming word and test word together without participants interpreting them as a compound \citep{gagne:2001}. 
This procedure isolates the experimental priming to each concept in the combination. 
For example, the lexical decision task applied to ``coil" and ``spring" was designed to prime the coil sense of the concept SPRING in the conceptual combination SPRING PLANT. 
The order of the two double lexical decision tasks was counter-balanced, so that half were presented in the same order as the compound words (e.g., ``coil" and ``spring" were first presented, then ``factory" and ``plant") and half were presented in the reverse order (e.g., first ``factory" and ``plant" were presented for lexical decision, followed by ``coil" and ``spring". \\
\noindent
{\bf Phase 3:}\\
 A conceptual combination was presented in the center of the screen (e.g., ``spring plant''). 
 Participants were asked to push the space bar as soon as they thought of an interpretation for the compound. Filler compounds were included for the filler (i.e., non-word) trials so as not to disrupt the participant's rhythm in making two lexical decisions followed by an interpretation. \\
\noindent
{\bf Phase 4:}\\
Participants were asked to type in a description of their interpretation.\\ 
\noindent
{\bf Phases 5-6:}\\
 Two disambiguation tasks were carried out, where participants chose what sense they gave to each word from a list (e.g., plant = A. `a living thing'; B. `a factory'; C. `other'). 

\subsection{Results}
Experimental subcomponents utilizing non-words were discarded during the  analysis. In total, 91.5\% of the interpretations provided by the subjects fell within one of the four primed senses of the studied conceptual combinations. 

As stated previously, in order to apply Fine's theorem or the CHSH inequalities for compositional analysis, marginal selectivity must first be tested.
Table \ref{tab:marginals} depicts an analysis of marginal selectivity where the values in the columns depict the difference of marginal probabilities across the conditions of the associated variable.
For example, diff($\rv{A1}$) is the difference between the one-marginal 
$\Pr(A1,B1)+ \Pr(A1,\bar{B1})$ and the other second marginal $\Pr(A1,B2)+ \Pr(A1,\bar{B2})$.
Marginal selectivity holds when these differences are zero across all for variables.
The statistical test reveals five conceptual combinations that fail marginal selectivity.
However, as there are usually sixteen data points in each of the four pair-wise distributions (see equation \eqref{pstruct}), the statistical test is underpowered.
As a consequence, there is uncertainty about which combinations are failing marginal selectivity. 
For the purposes of illustration, the four conceptual combinations BATTERY CHARGE, BILL SCALE, TABLE FILE and TOAST GAG are assumed to satisfy marginal selectivity, as these are the four combinations closest to satisfying marginal selectivity based on the sum of their Chi square values (see table \ref{tab:marginals}). 

The result of the compositional analysis in depicted in Table \ref{tab:results}.
Despite the findings of the analysis of marginal selectivity, we have adopted a conservative approach and flagged combinations as ``non-compositional" based solely on the assumption that they are likely to fail marginal selectivity.
Of the combinations that are assumed to satisfy marginal selectivity, BILL SCALE (|CHSH| =1.63) , TOAST GAG (|CHSH| =1.63) and TABLE FILE (|CHSH| = 0.33) are deemed ``compositional" as their CHSH values are less than or equal to 2.
BATTERY CHARGE shows a slight violation of the CHSH inequalities (|CHSH| = 2.01), but due to the lack of statistical power just mentioned, we cannot conclude whether this combination does represent an actual case where non-compositionality manifests in the presence of marginal selectivity being satisitfied. 
Only a subsequent study with sufficient statistical power can resolve whether such conceptual combinations exist.


\begin{table}[h]
\centering
\scalebox{0.75}{
\begin{tabular}{|l|l|l|l|l|}
\hline
Combination & diff ($\rv{A1}$)  & diff ($\rv{A2}$) & diff  ($\rv{B1}$) & diff  ($\rv{B2}$) \\\hline
boxer bat & 0.175 (0.46) &0.140  (0.24)  & 0.338 (2.6) & 0.158 (0.27)\\\hline
bank log & 0.055  (0) & 0.092 (0.02) & 0.338 (3.30*) & 0.257 (1.73)  \\\hline
apple chip* & 0.250 (2.77*) &  0.114 (0.09)  & 0.294 (2.88*) &  0.217 (0.78) \\\hline
stock tick* & 0.163 (0.30) & 0.085 (0.02)   & 0.488 (5.59*) & 0.386 (3.77*) \\\hline
seal pack & 0.083 (0.01) & 0.213 (0.77)  & 0.162 (0.38) & 0.221 (0.78) \\\hline
spring plant* & 0.294 (3.49*) & 0.133  (0.61) & 0  & 0.173 (0.81) \\\hline
poker spade & 0.136 (0.25)  & 0.035 (0) & 0 (0) & 0.113 (0.09) \\\hline
slug duck & 0.096 (0.03) & 0.153 (0.32) & 0.133 (0.21)  & 0.026 (0) \\\hline
club bar & 0.133 (0.68) & 0 (0) & 0.125  (0.60) & 0.138 (0.37) \\\hline
web  bug  & 0.210 (0.74) & 0.067 (0)  & 0.296 (1.70) & 0.153 (0.32) \\\hline
{\bf table file} & 0.058 (0)  & 0.235 (0) & 0.114 (0.09) & 0.113 (0.09) \\\hline
match bowl & 0.137 (0.18) & 0.250 (1.31) & 0.075 (0.01) & 0.022 (0) \\\hline
net cap & 0.035 (0) & 0.092 (0.03) & 0.059 (0) & 0.175 (0.46) \\\hline
stag yarn* & 0.375 (3.64*)  & 0.219 (1.14)  & 0.104 (0.05)  & 0.045 (0)\\\hline
mole pen & 0.125 (0.29) & 0.021 (0)  & 0.063 (0) & 0.3 (1.87)\\\hline
{\bf battery charge} & 0.067 (0)  & 0.048  (0) & 0.117 (0)  & 0.120 (0.08)\\\hline
count watch & 0.195 (0.89)  & 0.063 (0) & 0.011 (0) &0.063 (0)  \\\hline
{\bf bill scale} & 0.081 (0.02) & 0.113 (0.09) & 0.054 (0) & 0.051 (0) \\\hline
rock strike & 0.188 (1.47) & 0.117 (0.13)  & 0.313 (3.79)  & 0.013 (0)\\\hline
port vessel & 0.106 (0.07) & 0.085 (0.02)   & 0.113 (0.09) & 0.118  (0.20) \\\hline
crane hatch & 0.141 (0.45) & 0.296  (1.70) & 0.149 (0.39) & 0.233 (0.93)  \\\hline
{\bf toast gag} & 0.0625 (0) & 0.008 (0)   & 0.018 (0) & 0.015  (0) \\\hline
star suit & 0.308 (2.63) &0.163 (0.28) & 0.054 (0)  & 0.058 (0)  \\\hline
fan post & 0.35 (2.59)  &0.125 (0.13) & 0.025 (0)  & 0.188 (0.55)\\\hline
\end{tabular}
} 
\\
~\\
\caption{Analysis of marginal selectivity. Starred conceptual combinations fail marginal selectivity (Chi square test of proportions: critical value=2.71 ($\alpha= 0.1$). Bolded conceptual combinations look to satisfy marginal selectivity \label{tab:marginals}}
\end{table}
\begin{table}[h]
\scalebox{0.65}{
\begin{tabular}{|l|l|l|l|l|l|l|l|l}
\hline
 & Concept A &  & Concept B & &Results &  &\\\hline
Combination & Prime 1($\rv{A1}$)  & Prime 2 ($\rv{A2}$) & Prime 3  ($\rv{B1}$) & Prime 4 
($\rv{B2}$) & $Compositional$ &  $N$  \\\hline
boxer bat &\emph{dog}&\emph{fighter}&\emph{ball}&\emph{vampire}&   N  & 64  \\\hline
bank log &\emph{money}&\emph{river}&\emph{journal}&\emph{tree}&  N  &  64 \\\hline
apple chip &\emph{banana}&\emph{computer}&\emph{potato}&\emph{circuit}& N &  65  \\\hline
stock tick &\emph{shares}&\emph{cow}&\emph{mark}&\emph{flea} &   N  &64\\\hline
seal pack &\emph{walrus}&\emph{envelop}&\emph{leader}&\emph{suitcase}& N & 64\\\hline
spring plant &\emph{summer}&\emph{coil}&\emph{leaf}&\emph{factory}& N &64\\\hline
poker spade &\emph{card}&\emph{fire}&\emph{ace}&\emph{shovel}&  N   &65\\\hline
slug duck &\emph{snail}&\emph{punch}&\emph{quack}&\emph{dodge}&  N   &63\\\hline
club bar &\emph{member}&\emph{golf}&\emph{pub}&\emph{handle}&  N &64\\\hline
web  bug &\emph{spider}&\emph{internet}&\emph{beetle}&\emph{computer}&  N   &63\\\hline
{\bf table file} &\emph{chair}&\emph{chart}&\emph{nail}&\emph{folder}&    Y [0.33]   &63\\\hline
match bowl&\emph{flame}&\emph{contest}&\emph{disk}&\emph{throw}& N  &64\\\hline
net cap&\emph{gain}&\emph{volleyball}&\emph{limit}&\emph{hat}&  N   &65\\\hline
stag yarn&\emph{party}&\emph{deer}&\emph{story}&\emph{wool}&  N   &61\\\hline
mole pen&\emph{dig}&\emph{face}&\emph{pig}&\emph{ink}&   N      &63\\\hline
{\bf battery charge} &\emph{car}&\emph{assault}&\emph{volt}&\emph{prosecute}& ? [2.01]  &63\\\hline
count watch&\emph{number}&\emph{dracula}&\emph{time}&\emph{look}& Y     &65\\\hline
{\bf bill scale} &\emph{phone}&\emph{pelican}&\emph{weight}&\emph{fish}& Y [1.63]   &64\\\hline
rock strike &\emph{stone}&\emph{music}&\emph{hit}&\emph{union}&   N   &63\\\hline
port vessel&\emph{harbour}&\emph{wine}&\emph{ship}&\emph{bottle}&  N   &65\\\hline
crane hatch&\emph{lift}&\emph{bird}&\emph{door}&\emph{egg}& N      &63\\\hline
{\bf toast gag} &\emph{jam}&\emph{speech}&\emph{choke}&\emph{joke}&  Y [1.23]   &63\\\hline
star suit&\emph{moon}&\emph{movie}&\emph{vest}&\emph{law}&   N      &62\\\hline
fan post &\emph{football}&\emph{cool}&\emph{mail}&\emph{light}&  N  &63\\\hline
\end{tabular}
} 
\\
\caption{Results of the compositionality analysis:  `Y/N' indicates whether the conceptual combination is compositional, or not  $N$ the number of subjects. Conceptual combinations classified as compositional are bolded with associated CHSH values in brackets}. 
\label{tab:results}
\end{table}

\subsection{Discussion}
\label{sec:discussion}
In this discussion further details are provided in order to shed light on \emph{how}  the joint probability distribution is structured when a violation occurs which serves to illustrate a number of key features about non-compositionality.
In what follows, we shall utilize two  examples: TOAST GAG and APPLE CHIP.
\subsubsection*{TOAST GAG}
Table~\eqref{eq:toastgag} depicts the the empirical results for TOAST GAG. Here, we see no particular ordering or patterns. In particular, when we compare the form of the equation required for a violation \eqref{eq:cereceda} and the the actual values in table \eqref{eq:toastgag} we can see that the probability mass does not center sufficiently around the diagonals in such a way that it can produce the correlations between the senses necessary to violate the CHSH inequality as $|\text{CHSH}=1.23| \leq 2$. 
The conceptual combination TOAST GAG is therefore deemed to be ``compositional" as a joint probability distribution $\Pr(\rv{A1}, \rv{A2}, \rv{B1}, \rv{B2})$ can be constructed, which models how it is interpreted within the given priming conditions.
\begin{equation}
\label{eq:toastgag}
 \begin{array}{c}
\begin{sideways}\text{TOAST}\end{sideways}
\end{array}%
\hspace*{-0.2cm}
\begin{array}{c}
\underset{}{%
\begin{array}{c}
\rv{A1} (jam)%
\end{array}%
\begin{array}{r}
+1 \\ 
-1%
\end{array}%
} \\ 
\overset{}{%
\begin{array}{c}
\rv{A2} (speech)%
\end{array}%
\begin{array}{r}
+1 \\ 
-1%
\end{array}%
}%
\end{array}%
\overset{\overset{%
\begin{array}{c}
\text{GAG}%
\end{array}%
}{%
\begin{array}{cc}
\overset{%
\begin{array}{c}
\rv{B2} (choke)%
\end{array}%
}{%
\begin{array}{cc}
+1 &\quad -1%
\end{array}%
} & \overset{%
\begin{array}{c}
\rv{B1} (joke)%
\end{array}%
}{%
\begin{array}{cc}
+1 &\quad -1%
\end{array}%
}%
\end{array}%
}}{\left( 
\begin{tabular}{c|c}
$\underset{}{%
\begin{array}{cc}
0.50 & 0.4375\\ 
0.0625 & 0%
\end{array}%
}$ & $\underset{}{%
\begin{array}{cc}
0.625& 0.375\\ 
0 & 0%
\end{array}%
}$ \\ \hline
$\overset{}{%
\begin{array}{cc}
0.29 & 0 \\ 
0.29 & 0.42%
\end{array}%
}$ & $\overset{}{%
\begin{array}{cc}
0.07 & 0.21\\ 
0.57& 0.14%
\end{array}%
}$%
\end{tabular}%
\right) } 
\end{equation}

\subsubsection*{APPLE CHIP}
In contrast, APPLE CHIP leads to a joint distribution that has a  more interesting structure:
\begin{equation}
\label{eq:applechip}
\begin{array}{c}
\begin{sideways}\text{APPLE}\end{sideways}%
\end{array}%
\hspace*{-0.2cm}
\begin{array}{c}
\underset{}{%
\begin{array}{c}
\rv{A1} (banana)%
\end{array}%
\begin{array}{r}
+1 \\ 
-1%
\end{array}%
} \\ 
\overset{}{%
\begin{array}{c}
\rv{A2} (computer)%
\end{array}%
\begin{array}{r}
+1 \\ 
-1%
\end{array}%
}%
\end{array}%
\overset{\overset{%
\begin{array}{c}
\text{CHIP}%
\end{array}%
}{%
\begin{array}{cc}
\overset{%
\begin{array}{c}
\rv{B1} (potato)%
\end{array}%
}{%
\begin{array}{cc}
+1 &\quad -1%
\end{array}%
} & \overset{%
\begin{array}{c}
\rv{B2} (circuit)%
\end{array}%
}{%
\begin{array}{cc}
+1 &\quad -1%
\end{array}%
}%
\end{array}%
}}{\left( 
\begin{tabular}{c|c}
$\underset{}{%
\begin{array}{cc}
0.94 & 0.06\\ 
0 & 0%
\end{array}%
}$ & $\underset{}{%
\begin{array}{cc}
0 & 0.75\\ 
0.25 & 0%
\end{array}%
}$ \\ \hline
$\overset{}{%
\begin{array}{cc}
0 & 0.35 \\ 
0.65 & 0%
\end{array}%
}$ & $\overset{}{%
\begin{array}{cc}
0.47 & 0\\ 
0 & 0.53%
\end{array}%
}$%
\end{tabular}%
\right) } 
\end{equation}

It is clear from the values that APPLE CHIP fails marginal selectivity.
Therefore the joint probability distribution $\Pr(\rv{A1},\rv{A2},\rv{B1},\rv{B2})$ cannot be constructed from the four empirically collected pairwise joint probability distributions such that these four pairwise distributions depcted in table \eqref{eq:applechip} can be recovered by marginalising this four way joint distribution.
This conceptual combination is therefore deemed ``non-compositional". 

APPLE CHIP shows a strong pattern of correlation between the senses across the four priming conditions because the probabilities are concentrated on the diagonals or reverse diagonals. 
Thus, whenever a subject interprets APPLE as a fruit they tend to interpret CHIP in its FOOD sense. Conversely, if APPLE is interpreted as a `computer' then a CHIP is interpreted as an `electronic device'.
This structure was quite common in the conceptual combinations that were studied.
A second key factor is that a non-zero value has been returned by the ensemble of subjects for one off-diagonal case $p_{2}=\Pr( \rv{A1}=+1,\rv{B1}=-1) =0.06$ (see Table~\eqref{pstruct}) 
Even though the food sense of CHIP has been primed, atypical interpretations of the compound are produced, for example, ``apple's growth is controlled by an internal chip".
\citet{costello:keane:2000}  identify three categories of non-compositionality in novel conceptual combinations, and atypical instances are at the basis of one of these categories.
Some other non-compositional combinations similarly showed atypical interpretations. 
For example, BANK LOG also exhibits a strong correlation between the senses: When BANK is interpreted as a financial institution, LOG tends to be interpreted as a ``record".
Conversely, when BANK is interpreted in it's ``river" sense, LOG is interpreted as a ``piece of wood". 
However, there were atypical cases where the senses cross over which produces an off-diagonal probability e.g., ``a record of a bank of a river".

We hypothesise that one way for a conceptual combination to be deemed non-compositional when marginal selectivity is satisfied, is for the probability mass to be largely concentrated along diagonals together with off-diagonal elements with small probabilities. These small probabilities reflect the atypical interpretations. \citep{costello:keane:2000}.
It is interesting to note that BATTERY CHARGE, which is borderline between ``compositional" and ``non-compositional", does exhibit this structure. 
Based on this structure we generated a hypothetical example (N=400), with one hundred data points in each quadrant (See equation \eqref{eq:violation}). The Chi square values for the differences in marginal selectivity are (0.06,0,0,0) for $\rv{A1},\rv{A2},\rv{B1},\rv{B2}$ respectively.
Hence there are minor differences in marginal probabilities, where the differences are not significant at the 90\%  level. 
The probabilities in equation \eqref{eq:violation} yield an absolute CHSH value of 2.06, therefore this hypothetical conceptual combination can be classified as ``non-compositional".
The atypical interpretations, highlighted by the bolded probabilities, are what force the CHSH value to exceed the threshold of two.
It is this threshold that marks the border between compositionality and non-compositionality.
It is not surprising that the absolute CHSH value is only slightly above 2,  as by their very nature, atypical interpretations are infrequent.
.\begin{equation}
\label{eq:violation}
\begin{array}{c}
\begin{sideways}\text{A}\end{sideways}%
\end{array}%
\hspace*{-0.2cm}
\begin{array}{c}
\underset{}{%
\begin{array}{c}
\rv{A1} (prime a1)%
\end{array}%
\begin{array}{r}
+1 \\ 
-1%
\end{array}%
} \\ 
\overset{}{%
\begin{array}{c}
\rv{A2} (prime a2)%
\end{array}%
\begin{array}{r}
+1 \\ 
-1%
\end{array}%
}%
\end{array}%
\overset{\overset{%
\begin{array}{c}
\text{B}%
\end{array}%
}{%
\begin{array}{cc}
\overset{%
\begin{array}{c}
\rv{B1} (prime b1)%
\end{array}%
}{%
\begin{array}{cc}
+1 &\quad -1%
\end{array}%
} & \overset{%
\begin{array}{c}
\rv{B2} (prime b2)%
\end{array}%
}{%
\begin{array}{cc}
+1 &\quad -1%
\end{array}%
}%
\end{array}%
}}{\left( 
\begin{tabular}{c|c}
$\underset{}{%
\begin{array}{cc}
0.85 & {\bf 0.05}\\ 
0 & 0.10%
\end{array}%
}$ & $\underset{}{%
\begin{array}{cc}
0 & 0.92\\ 
0..08 & 0%
\end{array}%
}$ \\ \hline
$\overset{}{%
\begin{array}{cc}
0 &  0.06 \\ 
0.86 & {\bf 0.08}%
\end{array}%
}$ & $\overset{}{%
\begin{array}{cc}
0.07 & 0\\ 
0 & 0.93%
\end{array}%
}$%
\end{tabular}%
\right) } 
\end{equation}

\subsubsection*{Frequency of Interpretations}

The frequency of interpretations was analysed using Wilcoxon signed-rank tests. 
The results are summarised in table~\ref{tab:frequency-interpretation}.
\begin{figure}[ht]
\center
\begin{tabular}{|l|l|l|}
\hline 
         & Consistent & Inconsistent \\ \hline
Overall & 6.88 & 4.72 \\   \hline
Same Order & 3.20 & 2.32 \\ \hline
Reverse Order & 3.67 & 2.40 \\ \hline
\end{tabular}
\caption{Mean Number of Interpretations (Consistent or Inconsistent) with the Primes by Prime Order (Overall, Same Prime Order, Reverse Prime Order)}
\label{tab:frequency-interpretation}
\end{figure}

As expected, overall participants gave significantly more interpretations  that were consistent with the primes (mean = 6.88), than inconsistent with the primes (mean = 4.72), $z = 4.06, p< .0001$. 
This provides evidence that the primes were affecting the interpretations given in the correct direction. 
To analyse whether the order in which the primes were shown had an effect on number of interpretations, we divided the consistent and inconsistent interpretations into whether the priming words were in the same order or reverse order to that of the compound. 
No significant differences were found. 
Furthermore, the priming effect was still present within the priming order conditions. 
That is, when prime order was the same, participants gave significantly more consistent interpretations (mean = 3.20) than inconsistent interpretations (mean = 2.32), $z = 2.77,p  = .006$. 
Likewise, when prime order was reversed, participants again gave significantly more consistent interpretations (mean = 3.67) than inconsistent interpretations (mean = 2.40), 
$z = 3.34,p = .001$. 
Overall, these results provide strong evidence that the priming was effective, and that it is independent of priming order.
\subsubsection*{Response time}
The speed of producing an interpretation was analysed according to whether it was consistent or inconsistent with regards to the priming words, and whether this was affected by prime order. 
It was expected that if the priming was effective then interpretations that were inconsistent with the primes would be produced slower than interpretations that were consistent with the primes. As seen in table~\ref{tab:response-time}, the mean response times were in the correct direction. 
Since a number of participants did not give responses for all of the categories, the number of participants in the analysis was 51. 
The analysis showed no main effect of Interpretation ($p = 0.297$), Prime Order ($p = 0.718$), nor an Interpretation x Prime Order interaction ($p = 0.994$). 
One likely reason for the non-significant effects is the large variance in response times (range = 369ms to 10035 ms), thus making it difficult for the mean differences to reach significance. For this reason we feel that the frequency scores are more reliable measures, and importantly these showed significant effects of priming.
\begin{figure}
\begin{minipage}{0.50\linewidth}
\centering

\begin{tabular}{|l|l|l|}
\hline 
         & Consistent & Inconsistent \\ \hline
Overall & 3095.51 & 3288.99 \\   \hline
Same Order & 3066.47 & 3273.25 \\ \hline
Reverse Order & 3083.10 & 3299.40 \\ \hline
\end{tabular}\\\hspace*{0.5cm}
(a)
\end{minipage}
\begin{minipage}{0.50\linewidth}
\centering
\begin{tabular}{|l|l|l|}
\hline 
         & Consistent & Inconsistent \\ \hline
Overall & 3138.76 & 3352.58 \\   \hline
Same Order & 3155.78 & 3329.85 \\ \hline
Reverse Order & 3098.23 & 3274.58 \\ \hline
\end{tabular}
(b)
\end{minipage}
\caption{Mean Response Time for Producing Interpretations (Consistent or Inconsistent) with the Primes by Prime Order (Overall, Same Prime Order, Reverse Prime Order)(a) Mean response times (ms) before analysis (N = 65) (b) Mean response times (ms) used in ANOVA (N = 51)}
\label{tab:response-time}
\end{figure}
\subsubsection*{Compound familiarity}
One concern is that the evidence for non-compositionality found in this study may be a function of familiarity. In particular, highly familiar compounds would be expected to require less combinatorial processing as the combined meaning may simply be retrieved from long term memory. 
We consider this possibility unlikely due to the experimental procedure followed.
The fact that both words are ambiguous allows the priming procedure to shift participants into considering new combined meanings. 
For instance, while most participants (86\%) interpreted 
SPRING PLANT as ``a plant that grows in spring",  when primed with `coil' and `leaf', 3\% of participants gave the interpretation ``a springy plant". 
Thus these participants have arguably been influenced by priming towards generating a new meaning, even though a highly common meaning already exists. 
In fact, as previously mentioned for spring plant and other compounds the findings of non-compositionality seem to depend upon participants producing novel meanings for the compounds. This finding goes against the hypothesis that non-compositionality is driven entirely by the retrieval of pre-stored meanings.
To test whether familiarity is associated with non-compositionality, we obtained hit rates for each compound by typing each into google with quotes. 
This measure of familiarity has been used in previous studies (e.g., \citep{ramm:halford:2012,wisniewski:murphy:2005}. 
It was found that the novelty of compounds based upon hit rates ranged from 144 (STAG YARN) to 9,460,000 (BATTERY CHARGE). 
To reduce the large variance obtained in the hit rates we transformed the scores into logs of ten. If familiarity is driving the non-compositionality results it would be expected that CHSH scores would be positively correlated with google hit rates. To test this we calculated a Pearson R correlation. 
This showed a weak positive correlation between the two variables, though this was non-significant, $r = 0.21, p = .337$. Thus we did not find evidence for the hypothesis that the non-compositionality of compounds in this study is driven by familiarity. However, as there were only 24 compounds under study, we acknowledge that there may not have been enough power to derive a significant correlation. 

More generally, the primes are an experimentally pragmatic means to manipulate the manner in which context affects the interpretation applied to conceptual combinations, and so they need
only influence the interpretation, not determine it. 
The violations that do occur arise only with respect to the reported priming conditions, and may not occur in a different experimental context.

\section{Broader reflections on compositionality and non-compositionality}
\label{sec:broader}

\citet{costello:keane:2000} classify non-compositional conceptual combinations  into three  categories depending upon how their apparent non-compositionality arises. 
Firstly, some combinations are deemed non-compositional because of  \emph{emergent} properties, which generally arise from a meaning which is based on a subset of atypical instances. The aforementioned PET FISH example is placed in this category.
A second set of conceptual combinations are classified  non-compositional due to the manner in which the senses of the combining words are extended beyond their standard usage, to refer to instances outside the categories usually named by those words. 
Finally, some conceptual combinations are classified as non-compositional because they make use of cognitive processes such as metaphor, analogy or metonymy in their interpretation.
\citet{costello:keane:2000} use the conceptual combination SHOVEL BIRD to illustrate all three categories:
\begin{enumerate}
\item A ``shovel bird'' could be a bird with a flat beak for digging up food 
\item A ``shovel bird'' could be a bird that comes to eat worms when you dig in the garden 
\item	A ``shovel bird'' could be a plane that scoops up water from lakes to dump on fires 
\item A ``shovel bird'' could be a company logo stamped on the handle of a shovel 
\item A ``shovel bird'' could be someone allowed out of jail (free as a bird) as long as he works on a road crew
\end{enumerate}
They argue that (1) and (2) are examples of the first category because a bird with a flat beak is atypical, whereas (3) illustrates the second category because it extends the sense of both SHOVEL and BIRD beyond their normal usage. Finally (4) and (5) are put forward as examples of third category due to their metaphoric nature.
\citet{costello:keane:2000} detail how their constraint-based theory of conceptual combination specifically relates to each of these categories.
The framework presented in this paper, however,  models  the non-compositionality of SHOVEL BIRD irrespective of the category of non-compositionality involved.
For example, SHOVEL has the sense of being a tool, or being shaped like a shovel. The concept BIRD has three senses in the preceding example:  relating to an animal, a plane, and a prisoner.
Thus, the concept BIRD could be modeled as consisting of both a dominant ANIMAL ($\rv{A1}$) and a subordinate PLANE ($\rv{A2}$) sense, and then Bell/CH inequalities or the CHSH inequalities applied to test for the non-compositionality of each combination resulting from a combination of SHOVEL with BIRD. 

In addition, there is no requirement in the presented analytical framework that the concepts be homographs.
We require only that there be ambiguity caused by multiple possible interpretations of a concept, and this readily presents.  
A WordNet analysis of the noun-noun combinations used in the compositional models explored by \citet{mitchell:lapata:2010} reveals that the vast majority have more than one synset and hence more than one shade of meaning, and these may even be related (as was the case for the polysemous concept SHOVEL).
Ambiguity could also derive from relations. For example, the CARIN model assumes that relations apply to the modifier, so in ADOLESCENT DOCTOR 
(taken from \citep{gagne:2001}), an ambiguity arises between the competing relations in ``doctor FOR adolescents" and ``doctor IS adolescent". 
Both of these possibilities for the concept ADOLESCENT could be accessed through priming, and then probabilistically represented with their corresponding variables $\rv{A1}$ or $\rv{A2}$ (\citet{gagne:2001} provides an experimental procedure for priming relations).
DOCTOR is also ambiguous because it is polysemous, e.g., a medical doctor, or someone holding a PhD. 
Both of these possibilities could be modeled by the variables  $\rv{B1}$ and $\rv{B2}$. This example shows that the analytical framework presented here could be applied to the study of (non-)compositionality in conceptual combinations which have already been considered in the literature.

As the framework is general, and can be empirically tested, we argue that it has wide applicability for the analysis of conceptual combinations.
However, the determination of compositionality  that this analysis provides must take into account the priming conditions of the test, which empirically simulate the context (e.g., the discourse context) of the interpretation; there is no result without a supplied context (in this case the priming).
This is also the case in quantum physics;  a system may be deemed compositional in one measurement context, and not in another.
It is the theory for empirically testing a dividing line between compositionality and non-compositionality that this article contributes, not an adjudication of the ongoing debate on compositionality in conceptual representation.
One test is based on the violation of the CHSH inequalities and the other is based on the Bell/CH inequalities. 
Both of these tests are examples of the more general joint distribution criterion (JDC) proposed by \citep{dzhafarov:kujala:2012}.
The JDC  is decided by solving a linear programming problem of the form $\rv{MQ}=\rv{P},\rv{Q}\geq 0$. 
In the context of this article, the vector \rv{P} would comprise the sixteen probabilities depicted in \eqref{pstruct} 
and $\rv{Q}$ represents the global joint distribution $\Pr(\rv{A1},\rv{A2},\rv{B1},\rv{B2})$. 
\citet{dzhafarov:kujala:2012} prove that if marginal selectivity does not hold, then there is no solution for $\rv{Q}$.
If marginal selectivity holds and no distribution $\rv{Q}$ can be found, then the associated conceptual combination can be deemed non-compositional. The failure of the linear programming solution is analogous to a violation of the Bell/CH or CHSH inequalities.

Both the Bell/CH inequalities and the CHSH inequalities  require that marginal selectivity is satisfied.
There has been some confusion about the role of marginal selectivity when applying these formal mechanisms to cognitive phenomena \citep{dzhafarov:kujala:2014}.
For example, \citet{aerts:gabora:sozzo:2013} present an experiment to establish whether the concepts ANIMAL and ACTS are  ``entangled" in the expression ``The Animal Acts".
Placed within the framework presented in this paper, the goal of the experiment was to determine whether the conceptual combination ANIMAL ACTS is compositional, or not.
The authors employed the CHSH inequality and achieve a violation, meaning the combination is ``entangled", i.e., non-compositional.
However, a subsequent analysis of the experiment showed that marginal selectivity does not hold \citep{dzhafarov:kujala:2014}.
The associated formal analysis shows that it is inapplicable to employ the CHSH inequality, or the Bell/CH inequalities, when marginal selectivity does not hold.
Therefore, marginal selectivity should be checked before the CHSH inequalities or the Bell/CH inequalities are applied, as has been followed in the empirical illustration presented  in this paper.
This is crucially different to the situation in quantum physics where marginal selectivity seemingly always holds.


A parsimonious approach to modelling a conceptual combinations entails that a \emph{single} model can describe how it is being interpreted, namely a global joint probability distribution can be constructed from the four empirically collected pairwise joint distributions such that the empirically collected distributions can be recovered from the global distribution.
\citet{barros:2012} labels this fact ``contextuality"  because the inability to construct the joint distribution over the four variables is equivalent to the inability to assign values to the four variables that is consistent with all the experimentally observed marginal distributions.
This notion of contextuality provides some insight into non-compositionality as presented in this article.
Intuitively, if the way the conceptual combination is being interpreted varies sufficiently across the different priming conditions it will not be possible to provide a global model of the interpretations which is consistent with how the the interpretations are behaving with respect to the marginal distributions. 
We contend that in such cases the combination is non-compositional and moreover provides an empirically testable dividing line between compositionality and non-compositionality.
This view also holds that non-compositionality is a context-sensitive notion.
\citet{dzhafarov:kujala:2014a} have extended this notion of contextuality to allow its determination within the presence of marginal selecticity not being satisfied. This constitutes an important theoretical development as it is possible that marginal selectivity is often not satisfied for conceptual combinations, and perhaps also for cognitive phenomena more broadly.

It appears that historically George Boole considered the problem of the constraints involved when trying to construct a global distribution of three variables from pairwise joint distributions \citep{pitowsky:george}, however, it is the Russian mathematician Vorob'ev who discovered results equivalent to that of Fine's theorem.
As he was a contemporary of Kolmogorov, who axiomatized probability theory, Vorob'ev was apparently ignored \citep{khrennikov:2010}. 
Thus, it was quantum physics that became famous for demonstrating the impossibility  of modeling entangled systems in a single probability space.
In our opinion, this is but a quirk of the past, and \citet{dzhafarov:kujala:2012} have independently shown how such results appear in cognitive psychology.
The history just sketched, together with the fact that both the CHSH inequalities and Fine's theorem are based solely on conventional probability theory, opens the possibility to non-controversially apply them outside of quantum physics \citep{aerts:aerts:broekaert:gabora:2000,khrennikov:2010,aerts:sozzo:2011}.

\section{Conclusions}

This article departed from the assumption that conceptual combinations may not exclusively exhibit compositional semantics.
The very idea of a non-compositional semantics has been resisted in the literature spanning  cognitive science, philosophy and linguistics, probably because the ``principle of compositionality" has had such a significant track record of success over a long period.
It is, however, precisely the assumption that semantics must \emph{necessarily be} of a compositional form that has been regularly questioned in a wide range of literature. Despite this state of confusion, few analytical approaches have been proposed that are capable of demarcating the difference between the two forms of behavior. 
We have shown that it is possible to analyse the manner in which the semantics of a given conceptual combination might be considered as compositional, or non-compositional. Indeed, it is perhaps timely to remind the reader that we do not argue against compositional semantics \emph{per se}. 
Rather, we have tried in this article to shed light on the line at which it breaks down: 
We believe that both compositional and non-compositional analyses will be necessary in order to provide a full account of the semantics of language.

The semantics of concepts were modeled in terms of the different senses in which a concept may be understood, where a given sense corresponds to the interpretation attributed to a particular ambiguous concept.
These senses have a reliable intersubjective cognitive underpinning, as they were grounded in terms of human word association norm data, which was used to predict the  probability that a subject would attribute a particular sense to an ambiguous concept.
Utilising formal frameworks developed for analysing composite systems in quantum theory, we presented two methods that allow the semantics of conceptual combinations to be classified as ``compositional" or ``non-compositional".
This classification differs from previous research in two ways. 
Firstly, compositionality is not graded, e.g., ``weak" vs. ``strong" compositionality. 
Secondly, the declaration of compositionality, or non-compositionality, is not an absolute classification, but context sensitive.
An empirical study of twenty-four novel conceptual combinations illustrates how the classifications can be applied.
Important corollaries are:
\begin{itemize}
\item Conceptual combinations violating marginal selectivity cannot be modeled in a single probability space across the four variables modelling the respective interpretations of the constituent concepts. Such conceptual combinations are immediately ``non-compositional"
\item When marginal selectivity does hold, and the Bell/CH inequalities or the CHSH inequalities are not violated, then the semantics of the conceptual combination cannot be modeled in a four way joint probability distribution, the variables of which correspond to how the constituent concepts are being interpreted in their respective dominant and subordinate senses. Such conceptual combinations are ``compositional".
\item When marginal selectivity does hold, and any of the  Bell/CH inequalities or the CHSH inequalities \emph{are} violated, then the semantics of the conceptual combination cannot be modeled in a four way joint probability distribution.  Such conceptual combinations are ``non-compositional".
\end{itemize}
This result could have a marked impact in modeling cognitive phenomena more generally, as these phenomena are frequently assumed to be compositional, and no thought is given as to whether the phenomenon can be modeled within a given probability space that the modeler constructs in terms of random variables. It is simply assumed that it can.
Experiments from quantum physics show that for entangled systems no such model exists.

Finally, this article shows quantum theory is a fruitful source of new theoretical insights and tools for modeling conceptual semantics as it has  already provided for other areas of cognition \citep{bruza:busemeyer:gabora:2009,aerts:2009,khrennikov:2010,busemeyer:pothos:franco:trueblood:2011, busemeyer:bruza:2012}.

\subsection*{Acknowledgements}
This project was supported in part by the Australian Research Council
Discovery grants DP0773341 and DP1094974, and by the U.K. Engineering and Physical Sciences Research Council, grant number: EP/F014708/2.
Welcome support was also provided by the Marie Curie International Research Staff Exchange Scheme: Project 247590, ``QONTEXT - Quantum Contextual Information Access and Retrieval"). We thank Ehtibar Dzhafarov and Jerome Busemeyer for informative discussions.  Thanks also to Dr. Mark Chappell (Griffith University) for his assistance in running the experiments.

\bibliographystyle{elsarticle-harv}

\end{document}